\documentclass[10pt,twocolumn,letterpaper]{article}
\usepackage{bm}
\usepackage{iccv}
\usepackage{times}
\usepackage{booktabs}
\usepackage{graphicx}
\usepackage{amsmath}
\usepackage{amssymb}
\usepackage{multirow}
\usepackage[ruled,linesnumbered]{algorithm2e}

\usepackage[pagebackref,breaklinks,colorlinks]{hyperref}

\usepackage[capitalize]{cleveref}
\crefname{section}{Sec.}{Secs.}
\Crefname{section}{Section}{Sections}
\Crefname{table}{Table}{Tables}
\crefname{table}{Tab.}{Tabs.}

 \iccvfinalcopy 
\def\iccvPaperID{4896} 

\ificcvfinal\fi

\begin{document}

\title{Unified Adversarial Patch for  Cross-modal Attacks in the Physical World}

\author{
Xingxing Wei$^{1,2}$\thanks{Corresponding author}, \quad Yao Huang$^{1}$, \quad Yitong Sun$^{1}$, \quad Jie Yu$^{1}$ \\
$^{1}$ Institute of Artificial Intelligence, Beihang University, Beijing 100191, China\\
$^{2}$ Hangzhou Innovation Institute, Beihang University, Hangzhou 311228, China\\
\tt\small \{xxwei, y\_huang, yt\_sun, sy2106137\}@buaa.edu.cn
}

\maketitle
\ificcvfinal\thispagestyle{empty}\fi

\begin{abstract}
    Recently, physical adversarial attacks have been presented to evade DNNs-based object detectors. To ensure the security, many scenarios are simultaneously deployed with visible sensors and infrared sensors, leading to the failures of these single-modal physical attacks. To show the potential risks under such scenes, we propose a unified adversarial patch to perform cross-modal physical attacks, i.e., fooling visible and infrared object detectors at the same time via a single patch. Considering different imaging mechanisms of visible and infrared sensors, our work focuses on modeling the shapes of adversarial patches,  which can be captured in different modalities when they change. To this end, we design a novel boundary-limited shape optimization to achieve the compact and smooth shapes, and thus they can be easily implemented in the physical world. In addition, to balance the fooling degree between visible detector and infrared detector during the optimization process, we propose a score-aware iterative evaluation, which can guide the adversarial patch to iteratively reduce the predicted scores of the multi-modal sensors.
    We finally test our method against the one-stage detector: YOLOv3 and the two-stage detector: Faster RCNN. Results show that our unified patch achieves an Attack Success Rate (ASR) of 73.33\% and 69.17\%, respectively. More importantly, we verify the effective attacks in the physical world when visible and infrared sensors shoot the objects under various settings like different angles, distances, postures, and scenes. 
\end{abstract}

\section{Introduction}

Deep Neural Networks (DNNs) are vulnerable to adversarial
examples \cite{szegedy2013intriguing}, which pose a serious security threat to DNNs-based object detectors in the physical world\cite{xu2020adversarial,zhu2021fooling,zhu2022infrared}.   
 These physical attacks can help to evaluate the robustness of DNNs deployed in real-life systems, which have important practical values. Recently, RGB sensors and thermal infrared sensors have been simultaneously used in many safety-critical tasks such as security monitoring. When performing object detection tasks,  visible images could provide abundant information on the target's texture in the daytime, while infrared images could display the target's thermal distribution at night. Thus, combining visible images and infrared images together will result in a round-the-clock application. To evade object detectors under such multi-modal imaging scenarios, a necessary way is to develop a cross-modal physical attack to fool visible object detectors and infrared object detectors at the same time.

However, current physical attacks are limited in the single-modal domain. For example, some studies \cite{thys2019fooling, dong2022viewfool, wei2022simultaneously} hide from object detectors in the visible modality, and some studies \cite{zhu2021fooling,zhu2022infrared,wei2022hotcold} hide from object detectors in the infrared modality. Because visible sensors and infrared sensors have different imaging mechanisms, these single-modal physical attacks cannot simultaneously attack multi-modal object detectors. Specifically, the generated perturbations in the visible modality can not be captured by infrared sensors, and in turn, changing the object’s thermal radiation would not be reflected in the visible light domain. There also exist some cross-modal attacks in the digital world \cite{abdelfattah2021adversarial,tu2021exploring,DBLP:journals/corr/abs-2006-13192}, but they usually aim to modify the image pixels or point clouds after sensors' imaging, ignoring the different imaging mechanisms across sensors, and thus cannot transfer to physical world well. 

\begin{table}[t]
\caption{Various adversarial attacks in different settings. }
  \begin{center}
   \begin{tabular}{c|c|c}
    \hline
     & Digital world   & Physical world\\
    \hline
 \multirow{2}{*}{Single-modal}    &  \cite{wei2022sparse},\cite{carlini2017towards},\cite{wei2019sparse},\cite{goodfellow2014explaining},  & \cite{thys2019fooling}, \cite{sun2020towards} \cite{komkov2021advhat},\cite{dong2022viewfool}\\ 
 & \cite{kurakin2016adversarial},\cite{madry2017towards} \etc &  \cite{wei2022adversarial},\cite{Wei_2023_CVPR},\cite{zhu2022infrared} \etc \\   \hline
    Cross-modal  & \cite{abdelfattah2021adversarial},\cite{tu2021exploring}\etc & Ours \\
    \hline
    \end{tabular}
    \label{tab:robustness evaluation}
  \end{center}
  \vspace{-0.5cm}
\end{table}

Based on these discussions, this paper aims to design \textbf{a unified cross-modal attack in the physical world} to fill in the gap of this area, as illustrated  in Table \ref{tab:robustness evaluation}. Generally speaking, the lack of features that can operate in different modalities is the reason to limit the availability of cross-modal physical attacks. Inspired by \cite{chen2022shape} and \cite{wei2022hotcold}, we utilize adversarial patches \cite{DBLP:journals/corr/abs-1712-09665} and further optimize their shapes to perform attacks. Because patches' shapes are a universal attribute for different imaging mechanisms, their changes can be well captured by both visible sensors and infrared sensors, making it an appropriate form to perform cross-modal attacks. However, there still exist two challenges to meet our goal: (1) The current shape models are either heteromorphic \cite{chen2022shape} or change-limited \cite{wei2022hotcold}, leading to the difficulty of physical implementation or poor attacks because of finite searching space. So how to design a  flexible shape optimization method while keeping shapes' smoothness is the first challenge. (2) During the optimization for the unified physical attacks, the drop speed  of visible object detector and infrared object detector may be inconsistent, which will result in the successful evasion for one object detector but failed evasion for another object detector. Therefore, how to balance the performance of two object detectors from different domains is another challenge. 

    \begin{figure}[t]
    \begin{center}
       \includegraphics[width=\linewidth]{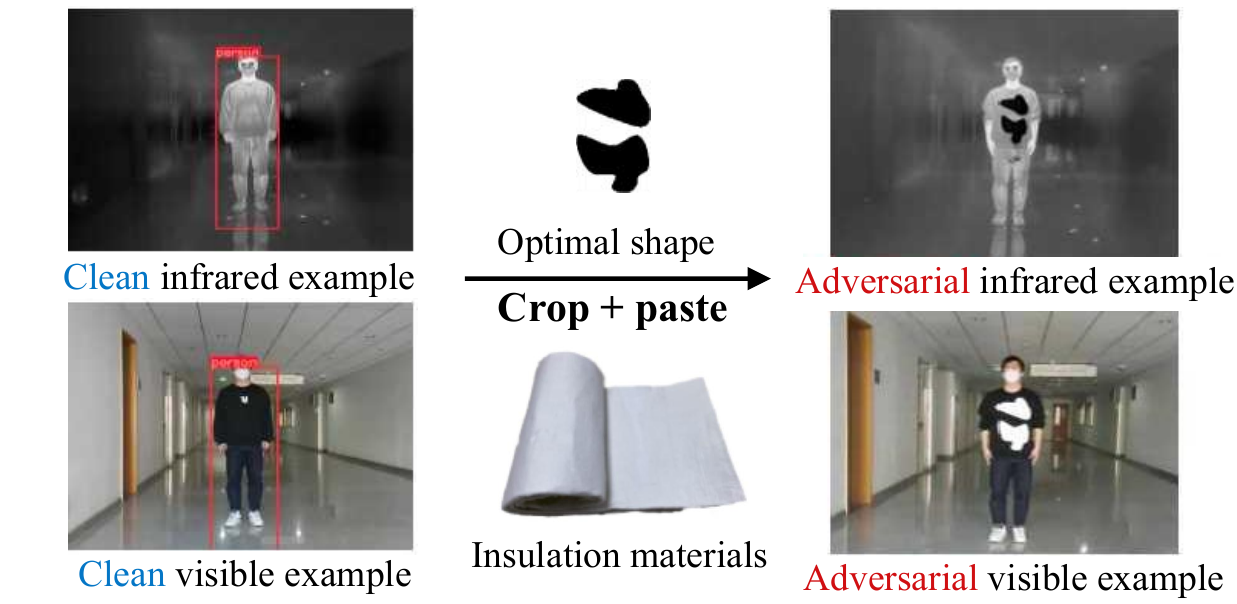}
    \end{center}
        \vspace{-0.3cm}
       \caption{The generation process for our unified cross-modal adversarial patches. We see the pedestrian cannot be detected after the patches are pasted on the pedestrian in the physical world.}
    \label{fig:pic1}
        \vspace{-0.3cm}
    \end{figure}

To meet the above challenges,   we design a novel boundary-limited shape optimization method to achieve the compact and smooth shapes, and thus they can be easily implemented in the physical world. Moreover, the shapes are flexible, and can provide a huge searching space to find the optimal shape to achieve a successful cross-modal attack.  In addition, to balance the fooling degree between visible detector and infrared detector during the optimization process, we propose a score-aware iterative evaluation, which can guide the adversarial patch to iteratively reduce the predicted scores of the multi-modal sensors. When applying to physical implements, we only need to print the simulated results in the digital world and crop them with insulation materials for patches. 
An example of our cross-modal adversarial patch against visible and infrared pedestrian detectors is shown in Figure \ref{fig:pic1}.

The contributions of this paper are as follows:
\begin{itemize}
     \item We propose a unified adversarial patch to perform cross-modal attacks. To the best of our knowledge, it is the first work to simultaneously evade visible detectors and infrared detectors in the physical world. 
     
    \item We design two novel techniques: boundary-limited shape optimization and score-aware iterative evaluation, to achieve the feasible patches in the digital world while balancing the multi-modal detectors.

    \item We verify our cross-modal patches in the pedestrian detection task not only in the digital world but also in the physical world. Experimental results show that cross-modal patches can work well in various angles, distances, postures, and scenes.
\end{itemize}

\begin{figure*}[ht]
\begin{center}
\includegraphics[width=0.9\linewidth]{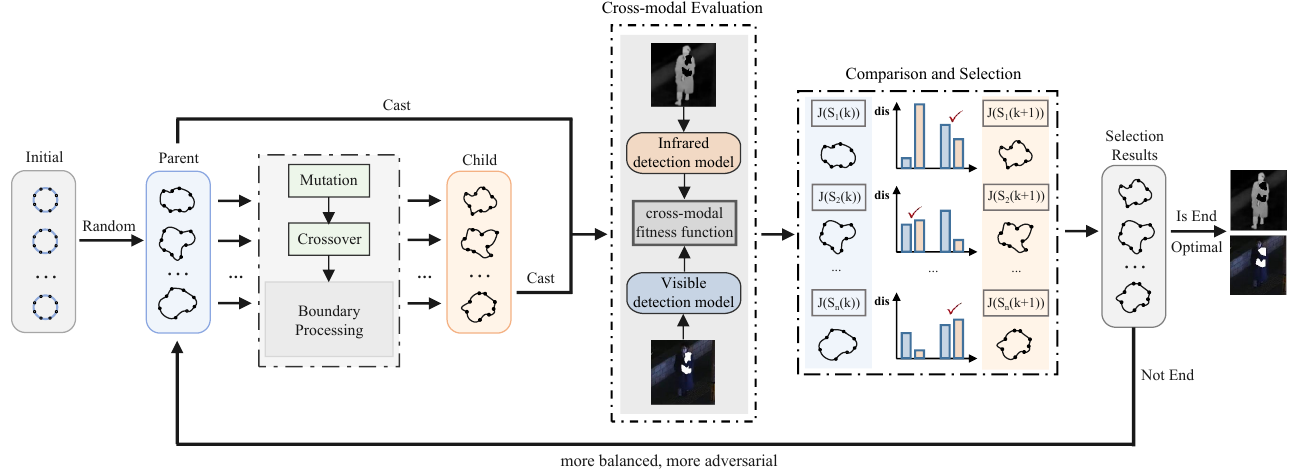}
\end{center}
   \caption{An overview of cross-modal adversarial patches' generation based on the Differential Evolution (DE) framework. The initial population is a series of circles. Then, combining mutation, crossover and boundary processing, a child population made up of natural shapes is generated. With a special cross-modal evaluation, we compare the parent population with the child population and select better individuals, encouraging the population to become more balanced and more adversarial. Finally, the optimal individual will be printed.}
\label{fig:framework}
\end{figure*}

\section{Related Works}

\subsection{Attacks in the physical world}
Since Kurakin \etal \cite{kurakin2016adversarial} verify the feasibility of adversarial attacks in the physical world, many physical attacks have been proposed for more application value, like Sharif \etal \cite{sharif2016accessorize}.'s adversarial glasses in face recognition systems, Eykholt \etal \cite{eykholt2018robust}'s adversarial graffiti in automatic driving tasks. Among them, 
adversarial patch \cite{brown2017adversarial} is one of the mainstream physical attacks.  Different from Lp-norm-based adversarial perturbations, the adversarial patch doesn't need to restrict the perturbations' magnitude, which is always in the form of a patch and thus more suitable to the physical world. Like Xu \etal \cite{xu2020adversarial}, they design a special adversarial T-shirt, which can apply the deformable adversarial patch on the T-shirts to fool the person detector.  When referring to infrared detectors, Zhu \etal propose adversarial bulbs \cite{zhu2021fooling} based on the extra heat source and invisible cloth \cite{zhu2022infrared} based on QR code pattern that could successfully attack infrared detectors, but compared with adversarial patches, they are complicated to implement in the physical world and lose effects in the visible modality.

\subsection{Shape model in adversarial attacks}

In addition to generating perturbations for the traditional adversarial patch, some studies explore the patches' shapes  to adversarial attacks. For example, Chen \etal \cite{chen2022shape} propose a deformable patch representation, but their deformation shapes mainly rely on an central point and rays, which causes the shape to be too sharp and not natural enough. In the meanwhile, their work still needs to optimize the contents on the patch to some extent and mainly focus on tasks of image classification. Besides, Wei \etal \cite{wei2022hotcold} propose a hotcold block based on Warming Paste and Colding Paste to attack infrared detectors, but  their deformation of patches is limited by the manually set nine-square-grid states, which greatly decreases the search space of patches' shapes. 

Overall, none of the above methods can simultaneously attack the two different modalities. Our work will combine adversarial patches with cross-modal attacks from the angle of a natural shape optimization.

\section{Methodology}
In this section, we choose the pedestrian detection as the target task to introduce the details of our method.
\subsection{Problem Formulation}

In the pedestrian detection task, given a clean visible image $x_{vis}$ and a clean infrared image $x_{inf}$, the goal of a unified cross-modal adversarial attack is to make the visible detector $f_{vis}(\cdot)$ and infrared detector $f_{inf}(\cdot)$ 
simultaneously unable to detect the pedestrian in the perturbed visible image $x^{adv}_{vis}$ and perturbed infrared image $x^{adv}_{inf}$. The formulation can be expressed as follows:
\begin{equation}
    max(f_{vis}(x^{adv}_{vis}), f_{inf}(x^{adv}_{inf})) < thre
    \label{attack}
\end{equation}
where $f_{vis}(x^{adv}_{vis})$ and $f_{inf}(x^{adv}_{inf})$ represent the confidence scores of detected pedestrians in the visible modality and infrared modality, and $thre$ is a pre-defined threshold.

The perturbed visible image $x^{adv}_{vis}$ and the perturbed infrared image $x^{adv}_{inf}$ with the unified adversarial patch can be generated as  Eq.(\ref{eq:x_vis_adv}) and Eq.(\ref{eq:x_inf_adv}). 
\begin{equation} 
    x^{adv}_{vis} = x_{vis}\odot (1 - M) + \hat{x}_{vis}\odot M
    \label{eq:x_vis_adv},
\end{equation}
\begin{equation}
    x^{adv}_{inf} = x_{inf}\odot (1 - M) + \hat{x}_{inf}\odot M
    \label{eq:x_inf_adv},
\end{equation}
where $\odot$ is Hadamard product, $M \in \{0,1\}^{h\times w}$ is a mask matrix used to constrain the shape and location of the cross-modal patches on the target object, $\hat{x}_{vis}\in R^{h\times w}$ denotes a cover image used to manipulate $x^{adv}_{vis}$, $\hat{x}_{inf}\in R^{h\times w}$ denotes a cover image used to manipulate $x^{adv}_{inf}$. The values of these two cover images are obtained by the visible sensor and infrared sensor versus the unified insulation material in the physical world. The unified patches can be described as all the regions where $M_{ij} = 1$. 

In the real application, we use the aerogel material to implement our unified adversarial patch. When infrared sensor shoots, it will show good insulation effects, and thus can change the thermal distribution of the target pedestrian. When visible sensor shoots, its color is white and can show a difference with the pedestrian. Based on this, our method  optimizes a cross-modal $M$ to learn an adversarial shape, finally working well in both modalities. 

\subsection{Shape Representation}
To model a shape, we need to determine the shape representation. For that, we first define some points as the basic elements. Then we connect these anchor points to construct the contour, representing our shape. Here are the details.

\textbf{Multi-anchor Representation:}
Unlike Chen \etal \cite{chen2022shape} using a central point and corresponding rays to form a polygon contour, we only use points, distributing multiple anchor points on the patch contour. Then, we can directly adapt the patch contour's shape by changing the coordinates of the points. Owing to the design of multi-anchor representation, we can have more flexible shape variations and a broader search space.

\begin{figure}[ht]
\begin{center}
\includegraphics[width=\linewidth]{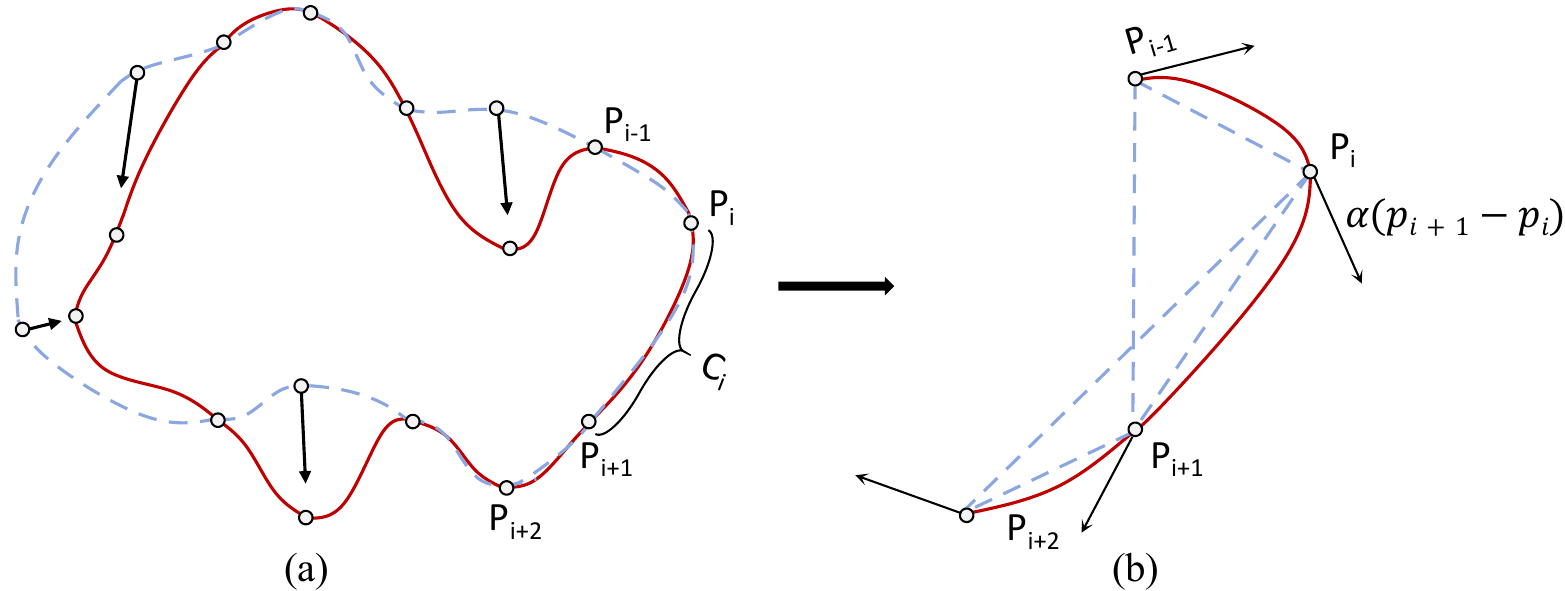}
\end{center}
\caption{Subfigure (a) shows the process of moving the anchor points to change the shape, where the dotted line denotes the initialized location, and arrows show the changed direction. Subfigure (b) is an example of the curve segment $C_{i}$'s connection via Catmull-Rom spline interpolation. }
\label{fig:patch_contour}
\vspace{-0.3cm}
\end{figure}

We illustrate this process in Figure \ref{fig:patch_contour} (a), where points in the dotted line denote the initialized location. The arrows show the changed direction for some points. We can adjust the coordinates of the points to control the movement of each point and thus, the direction of our points' movement will not be limited to one single direction.

\textbf{Smooth Spline Connection:}
To ensure shapes' naturalness, we use centripetal Catmull–Rom spline \cite{twigg2003catmull} to connect such anchor points. The process of spline can be formulated as follows:

As shown in Figure \ref{fig:patch_contour} (b), to generate a curve segment $C_i$ between $P_i$ and $P_{i+1}$, we will use four anchor points $P_{i-1},P_{i},P_{i+1},P_{i+2}$ and a centripetal Catmull-Rom spline function $CCRS(\cdot)$ \cite{twigg2003catmull}. With the centripetal Catmull-Rom spline function, first, we can ensure that a loop or self-intersection doesn't exist within a curve segment. Second, a cusp will never occur within a curve segment. Finally, it can also follow the anchor points more tightly. $C_i$ can be formulated as follows:
\begin{equation}
    C_{i} = CCRS(P_{i-1},P_{i},P_{i+1},P_{i+2})
\label{C_i}
\end{equation}

When $n$ curve segments $C_0,C_1,\cdots,C_{n-1}$ are combined, the patch contour $M_{con}$ can be written as:
\begin{equation}
    M_{con} = \{C_i|0\leq i\leq n-1\}
    \label{Mcon}
\end{equation}

With $M_{con}$ being closed, we can easily obtain $M$:
\begin{equation}
M(x)= \begin{cases}
1,&\text{$x$ \textit{inside} $M_{con}$} \\
0,&\text{$x$ \textit{out of} $M_{con}$}
\end{cases} 
    \label{M}
\end{equation}
After Eq.(\ref{M}), we can represent the patch's shape. The detailed formula for this process can be found in \textit{Supplementary Material}.
\label{deformable}
\begin{figure*}[ht]
\begin{center}
\includegraphics[width=\linewidth]{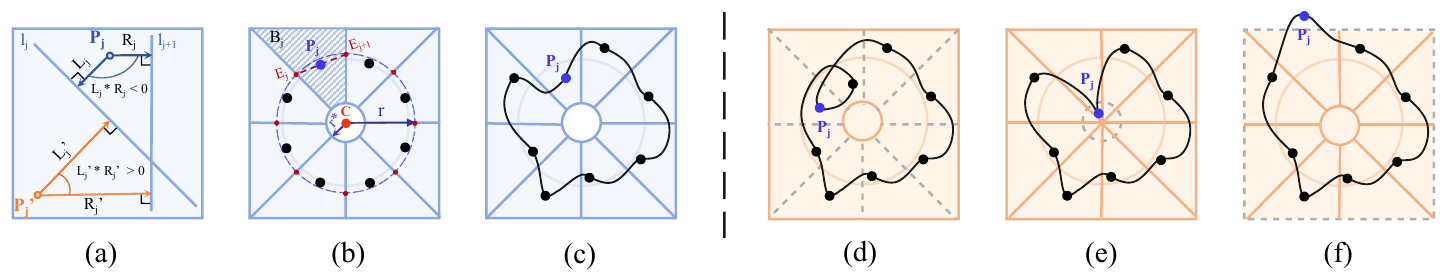}
\end{center}
   \caption{The solid blue and orange box represents the effective area of the human body and the dashed line means that the part is missing. The (a) expresses that the product of the directed distances from the point in the sharp-angled sector to the two side lines is less than 0 and if not, the product is more than 0. The (b) is our initial state and the (c) is an effective deformation. The (d) shows the patch contour will cross if we don't fix its relative order. The (e) represents that connecting points will be trapped into a restricted space if we don't set an inner circle. The (f) expresses that the patch contour will go beyond the human body if we don't define an outer boundary.}
\label{fig:boundary}
\end{figure*}
\subsection{Shape Optimization for Cross-modal Attacks}

In practice, the details of $f_{vis}(\cdot)$ and $f_{inf}(\cdot)$ are usually unknown to the adversary, so it is intractable to optimize the anchor points via a gradient-based optimization method. Considering this background, we carry out a score-based black-box attack by querying the object detector to obtain the confidence scores of detected pedestrians. From the above, we formulate the whole unified cross-modal attacks into the Differential Evolution (DE) framework.

\subsubsection{Formulation Overview Using DE}
The Differential Evolution (DE) consists of four parts: starting from an initial  population, using the crossover and mutation to generate the offspring population, making the fittest survive according to the fitness function, and finding the appropriate solution in the iterative evolution process.

In our case, a population represents the anchor points $\{P_j|j=1,...,n\}$. Given the population size $Q$, the $k$-th generation solutions $S(k)$ is represented as:
\begin{equation}
    S(k)=\{S_{i}(k)|\theta^{L}_{j}\leq S_{ij}(k)\leq \theta_{j}^{U}, 1\leq i\leq Q, 1\leq j\leq n\}
    \label{S(k)}
\end{equation}
where $S_{i}(k)$ is the $i$-th patch's shape, and $S_{ij}(k)$ represents the $j$-th anchor point of $S_{i}(k)$ in the $k$-th generation. $\theta^{L}_{j}$ and $\theta^{U}_{j}$ together make up the feasible region $B_{j}$, which is the moving range of the $j$-th anchor point in each patch shape.

In the $k+1$ generation of DE, the solution $S(k+1)$ is achieved via crossover, mutation, and selection based on $S(k)$. A fitness function is applied on $S_{i}(k)$ to evaluate its attack effectiveness. During this process, the fitness function only utilizes the confidence scores of object detectors.  We will give the detailed definition  of feasible region in Section \ref{boundary}, and fitness function in Section \ref{fitness}.

\subsubsection{Boundary-limited Shape Optimization}
\label{boundary}

Obviously, if there is no boundary restriction for the anchor points during the crossover and mutation,  some bad situations will occur during shape optimization, as shown in Figure \ref{fig:boundary} (d, e, f),  such as boundary line crossings (d), anchor points stuck in narrow places (e), and anchor points outside the human body region (f). To address these issues, we propose a novel boundary-limited shape optimization, which not only ensures the effectiveness of each shape, but also provides sufficient transformation space. 

As shown in Figure \ref{fig:boundary} (b), the shaded part $B_{j}$ is exactly the feasible region of the anchor point $P_{j}$. Specifically, the two adjacent equidistant lines, the inner circle's edge and the outer border together make up of our boundary. Figure \ref{fig:boundary} (c) presents us with an example of effective deformation. Next, we will give how to construct the boundary and the method of judging whether a point is inside or not.

We first divide a circle with a given radius $r$ and a circle center $C$ into $n$ sharp-angled sectors by $n$ equidistant lines $\{l_{j}|j=1,\cdots,n\}$. $n$ equal points $\{E_{j}|j=1,\cdots,n\}$ are the intersection of equidistant lines and the circle, and as shown in Figure \ref{fig:boundary} (b), the anchor point $P_{j}$ is the midpoint of $E_{j}$ and $E_{j+1}$, which can be formulated as follows:
\begin{equation}
    P_{j} = \frac{E_{j}+E_{j+1}}{2},\quad j=1,\cdots,n
\label{equal point}
\end{equation}

For the anchor point $P_{j}$, it lies in the sharp-angled sector region $B_{j}$ wrapped by two adjacent lines $l_{j}$ and $l_{j+1}$. $L_{j}$ and $R_{j}$ represents the directed distance from the anchor point $P_{j}$ to the line $l_{j}$ and $l_{j+1}$ shown in the Figure \ref{fig:boundary} (a). Therefore, with the line function $l_{j}(x,y)=0$ and $l_{j+1}(x,y)=0$, we can formulate $L_{j}$ and $R_{j}$ as follows:
\begin{equation}
    L_{j} = \frac{l_{j}(x_j,y_j)}{D_{j}},\quad R_{j} = \frac{l_{j+1}(x_j,y_{j})}{D_{j+1}}
\end{equation}
where $D_{j}$ and $D_{j+1}$ are the denominators of the point-to-line distance formula.
 
From the Figure \ref{fig:boundary} (b), we can vividly find that since the anchor point $P_{j}$'s feasible region is a sharp-angled sector, if $P_{j}$ is inside its own region, $L_{j}*R_{j}<0$ for the angle between them is obtuse and if $P_{j}$ is outside its own region or exactly on the boundary line,  $L_{j}*R_{j}\geq0$. With $P_{j}$ inside its own region, we can effectively avoid boundary line crossings in Figure \ref{fig:boundary} (d) caused by spline interpolation's need for a given order. 

Then, to prevent points from falling into narrow places in Figure \ref{fig:boundary} (e), we set an inner circle with a given radius $r^{*}$ inside the initial circle, and anchor points are not allowed to move into the inner circle. Specifically, we can judge by the distance $r_{j}$ from the anchor point $P_j$ to the center $C$. When $r_{j}>r^{*}$, the anchor point $P_{j}$ will not be inside the inner circle. The $r_{j}$ is computed as follows:  
\begin{equation}
    r_j = \sqrt{(x_{j}-x_{C})^2+(y_{j}-y_{C})^2}
\end{equation}

After that, we still need to ensure that the generated patch is not outside the effective area of the human body as shown in Figure \ref{fig:boundary} (f), so we scale the detection box output in a certain proportion as the outer border. The region $\mathcal{O}$ inside the outer border can be represented as follows:
\begin{equation}
    \mathcal{O} = \{(x,y)|x_{l}\leq x\leq x_{r}, y_{d}\leq x\leq y_{u}\}
\end{equation}
where $x_{l},x_{r},y_{u},y_{d}$ are outer border’s vertex coordinates.

Finally, we combine the above limits as follows:
\begin{equation}
    \rho_{j} = (L_{j}*R_{j}<0)\ \&\ (r_{j}>r^{*})\ \&\ (P_{j}\in \mathcal{O})
\label{rho}
\end{equation}
where $\rho_{j} = 1$ means that the anchor point $P_{j}$ is inside the feasible region $B_{j}$ and $\rho_{j} = 0$ means that the anchor point $P_{j}$ is outside the boundary.

\subsubsection{Score-aware Iterative Evaluation}
\label{fitness}

In cross-modal attacks, a good attack effect of a single modality is ineffective, while it is common to have unbalanced attack effects in the two modalities. If this situation is not improved, it may give the attacker false signals about the progress of the attack. Thus, to balance the fooling degree between visible detector and infrared detector during the optimization process, we propose a score-aware iterative evaluation, guiding the adversarial patch to iteratively reduce predicted scores of the multi-modal sensors.

For the sake of simplicity of expression, here we denote $S_{ij}(k)$ as $s$. To evaluate the fitness value of $s$, we first use methods in Section \ref{deformable} to transform $s$ into a patch mask $M$, then $x^{adv}_{vis}$, $x^{adv}_{inf}$ are produced based on Eq.(\ref{eq:x_vis_adv}), Eq.(\ref{eq:x_inf_adv}). $J(s)$ can be formed as follows:
\begin{equation}
    J(s)=e^{\lambda*min(dis(x^{adv}_{vis}),dis(x^{adv}_{inf}))}
        \label{J(s)}
\end{equation}
where $\lambda$ is a weighted factor, $dis(x^{adv}_{vis}),dis(x^{adv}_{inf})$ reflect the current progress towards the success of attack
(the larger, the closer to success). $dis(x^{adv}_{vis}),dis(x^{adv}_{inf})$ can be formalized as:
\begin{equation}
    dis(x^{adv}_{vis})=\frac{f_{vis}(x_{vis})-f_{vis}(x^{adv}_{vis})}{f_{vis}(x_{vis})-thre}
        \label{dis_vis}
\end{equation}
\begin{equation}
    dis(x^{adv}_{inf})=\frac{f_{inf}(x_{inf})-f_{inf}(x^{adv}_{inf})}{f_{inf}(x_{inf})-thre}
        \label{dis_inf}
\end{equation}
where $f_{vis}(x_{vis})$ is the confidence score of $x_{vis}$ in the visible pedestrian detector, $f_{inf}(x_{inf})$ is the confidence score of $x_{inf}$ in the infrared pedestrian detector. $f_{vis}(x^{adv}_{vis})$ and $f_{inf}(x^{adv}_{inf})$ are similar to $f_{vis}(x_{vis})$ and $f_{inf}(x_{inf})$.

From Eq.(\ref{dis_vis}) and Eq.(\ref{dis_inf}), we can easily know $dis(x^{adv}_{vis})$ and $dis(x^{adv}_{inf})$ measure the progress to success of the cross-modal patch attack in the visible and infrared modality respectively. Both of them can help patches evolve in the corresponding modality. However, if we only make use of a single one, it will be certain to cause unbalanced phenomena, not an effective cross-modal attack. To solve this issue, we use Eq.(\ref{J(s)}) to combine $dis(x^{adv}_{vis})$ and $dis(x^{adv}_{inf})$. Based on our $J(\cdot)$, the good performance only in a single modality will not achieve a high value of fitness because we take the worse one of $dis(x^{adv}_{vis})$ and $dis(x^{adv}_{inf})$ as a standard. Additionally, considering the difference of attack difficulty in the initial stage and later stage, we use $e^{(\cdot)}$ instead of a linear function. Based on the above settings, we will eventually encourage it to iteratively evolve in the direction of reducing confidence scores as much as possible while maintaining the balance of cross-modalities.

The overall algorithm for generating cross-modal adversarial patches is summarized in Algorithm 1.

\begin{algorithm}[h]  
    \caption{\small Shape-based Cross-modal Attack} 
        \KwIn{Clean visible image $\bm{x}_{vis}$, clean infrared image $\bm{x}_{inf}$, fitness function $J(\cdot)$, population size $Q$, max iteration number $T$}

        \small Initialize a collection of shapes $S(0)$

        \tcc{Optimize patch shape with DE}
        
        \For{$k = 0$ to $T-1$}{
        
            \small Sort $\bm{S}(k)$ in descending order based on $J(S(k))$ 
            
            \If{$\bm{S_0(k)}$ makes the attack successful}{
            
            \small $stop=k$; break; 
            
            }
            
            \small Generate $S(k+1)$ based on crossover and  mutation
            
            \small Limit boundaries of $S(k+1)$ according to Eq.(\ref{rho})
            
            \For{$i=1$ to $Q$}{
            
            \small Evaluate $S_{i}(k)$, $S_{i}(k+1)$ according to Eq.(\ref{J(s)})
            
            \small{$S_{i}(k+1)\leftarrow$ the better one in $S_{i}(k)$, $S_{i}(k+1)$}
            
            }  
        
        }
        
        \small{Sort $\bm{S}(stop)$ in descending order according to $J(\cdot)$} 
        
        \tcc{Model patch shape from points}
        
        \small{Anchor points$\{P_{i=1,\cdots,n}\}\leftarrow S_{0}(stop)$}
        
        \small{Curve segments $\{C_{i=1,\cdots,n}\}\leftarrow$ connect $\{P_{i=1,\cdots,n}\}$ according to Eq.(\ref{C_i})}
        
        \small{$M_{con}\leftarrow$ combine $\{C_{i=1,\cdots,n}\}$ according to Eq.(\ref{Mcon})}
        
        \small{$M\leftarrow$ fill the $M_{con}$ according to Eq.(\ref{M})}
        
        \small{$x^{adv}_{vis},x^{adv}_{vis}\leftarrow$ according to Eq.(\ref{eq:x_vis_adv}),Eq.(\ref{eq:x_inf_adv})}
        
        \small{\small\textbf{return} $M$, $x^{adv}_{vis}$, $x^{adv}_{inf}$}
        
        \KwOut{Mask $M$, adversarial visible example $x^{adv}_{vis}$ and adversarial infrared example $x^{adv}_{inf}$}
\end{algorithm}

\subsection{Physical Implementation}
\label{physical}
 
\begin{figure}[ht]
\begin{center}
\includegraphics[width=\linewidth]{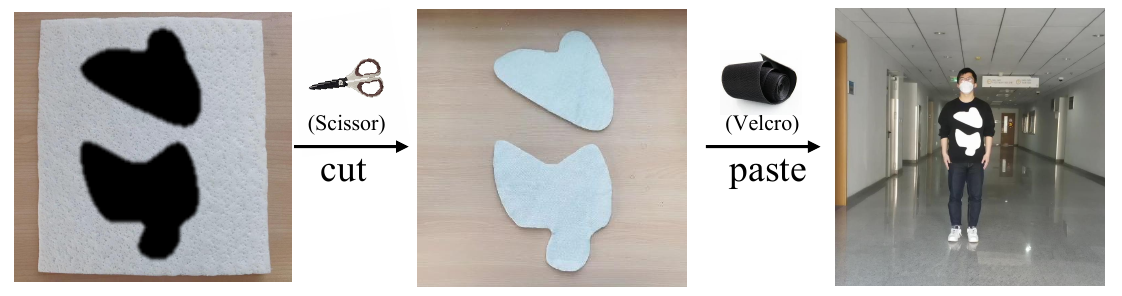}
\end{center}
  \caption{The physical implementation process of transforming digital cross-modal patches into physical patches in the real world.}
\label{fig:implementation}
\vspace{-0.3cm}
\end{figure}

After obtaining the optimal shape, we start to transform algorithm-generated digital patches into physical patches with the aerogel material. Specifically, we first scale the obtained patch according to its size in the real world and print it. Then, we use scissors to equally cut out the optimal shape on the material. And in the last, we use the velcro sticker to fix it in the corresponding position of the human body for performing a physical attack. The whole process is demonstrated in Figure \ref{fig:implementation}.

\section{Experiments}
\subsection{Simulation of Physical Attacks}
\noindent\textbf{Dataset:} We use the LLVIP \cite{jia2021llvip} dataset to simulate the physical attacks.  Images from LLVIP are perfectly synchronized in both visible modality and infrared modality. Similar to \cite{zhu2021fooling} and \cite{zhu2022infrared}, we customize the parts containing pedestrians from images in LLVIP. There are 1220 photos in the test set and 3784 images in the training set. As the final samples to be attacked, we pick 120 images from the test set that the target model can recognize successfully with a high probability. The clean AP is therefore 100\%. 

\noindent\textbf{Target detector:} For the pedestrian detection task, we select two mainstream detectors: YOLOv3 \cite{redmon2018yolov3}(one-stage) and Faster RCNN \cite{ren2015faster}(two-stage) here. About model training, we choose the officially pre-trained weights as the initialized weights and then retrain the model on the training dataset.
Results of some other detectors like  YOLOv5, YOLOv7 \cite{https://doi.org/10.48550/arxiv.2207.02696}, SSD \cite{liu2016ssd}, and EfficientDet\cite{tan2020efficientdet} are shown in the \textit{Supplementary Material}. 

\noindent{\textbf{Metrics}}: Attack Success Rate (ASR) and Average Precision  drop (AP drop) are used to evaluate the attack performance. Here, we adopt a unique cross-modal ASR, which denotes the ratio of both successfully attacked images under two modalities out of all the test images, highlighting the effectiveness of simultaneously attacking two modalities. AP drop is to show the AP's variation  before and after attacks.

\noindent\textbf{Implementation:}  As we use the DE algorithm, we set the number of the initial population as $30$, the epoches of evolution as $200$. Other parameters, formulas and comparative experiments are presented in the \textit{Supplementary Material}. 

\subsubsection{Performances in Different Detection Systems}
We first evaluate the effectiveness of attacks in the digital world. Considering the differences between detection systems, we choose two typical detectors: YOLOv3 and Faster RCNN. The results are shown in Table \ref{different detectors}. We use ASR and AP drop to evaluate the attack performance.

\begin{table}[!htb]
\centering
   \caption{Attack performances in different detection systems.}
\begin{tabular}{c |c| c}  \hline  
        & YOLOv3 & Faster RCNN   \\   \hline  
ASR     &    73.33\%    &           69.17\%                  \\   \hline  
AP drop (Visible) &   99.19\%   &       89.38\%                   \\     
AP drop (Infrared) &     74.31\%   &         83.94\%             \\   \hline  
\end{tabular} 
\vspace{-0.2cm}
\label{different detectors}
\end{table}
From the above results, we can see that our method is equally        
 useful despite the distinctions between the one-stage and two-stage detection models. For the typical one-stage detection model YOLOv3,  we achieve an ASR of 73.33\%, an AP drop of 99.19\% in the visible modality and 74.31\% in the infrared modality. Similarly, for the two-stage detection model Faster RCNN, we achieve an ASR of 69.17\%, an AP drop of 89.38\% in the visible modality and 83.94\% in the infrared modality.
 
\begin{figure}[ht]
\begin{center}
\includegraphics[width=\linewidth]{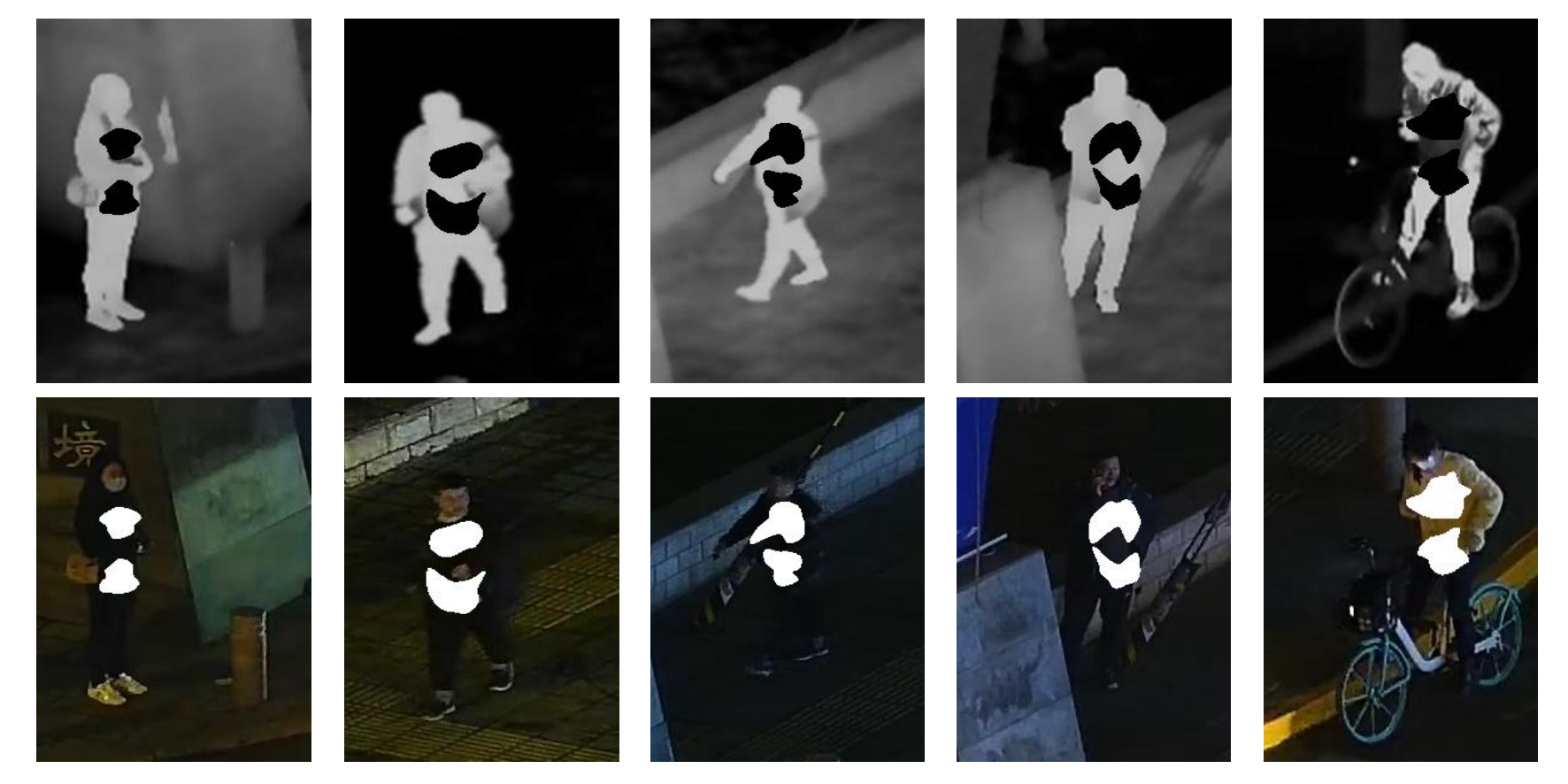}
\end{center}
\vspace{-0.5cm}
   \caption{Visual examples of adversarial samples with cross-modal patches in the digital world.}
\label{fig:digital_attack}
\vspace{-0.5cm}
\end{figure}

\subsubsection{Effects of Optimized Shapes}
\label{optimized shapes}
Here, we provide the ablation study to investigate the outcomes of our optimized shapes. Specifically, we first generate cross-modal adversarial patches to achieve the optimal shapes for a given pedestrian. Some examples of the optimal shapes in the digital world are shown in Figure \ref{fig:digital_attack}. Then, we fix the patches' locations and sizes while changing their shapes. These shapes can be circles, squares, rectangles and triangles, etc. Then we compute the ASR and AP drop under these five shapes. This setting tests the effects of optimal shapes on the attacks. The results are listed in Table \ref{shapes}, where we can see that random shapes barely work with an average ASR of 5.23\%, an average AP drop of 20.73\% in the visible modality and 11.65\% in the infrared modality. The results for each of the five shapes and compared with \cite{wei2022hotcold} are shown in the \textit{Supplementary Material}.
\begin{table}[h]\centering
\vspace{-0.2cm}
\caption{Ablation study for unified adversarial patches' shapes.}
\begin{tabular}{c | c | c }
\hline
        & Our shape & Other shapes\\ \hline
ASR     &    \textbf{73.33\%}  &   5.23\%      \\   \hline  
AP drop (Visible) &     \textbf{99.19\%} &   20.73\%  \\   
AP drop (Infrared) & \textbf{74.31\%} & 11.65\% \\  \hline  

\end{tabular}
\vspace{-0.5cm}
\label{shapes}
\end{table}

\subsubsection{Effects of Score-aware Iterative  Fitness Function}
\label{exp in cross-modal fitness function}
To verify the impact of the Cross-modal Fitness Function, we design a set of comparative experiments. As mentioned in Section \ref{fitness}, our method can combine the visible modality and infrared modality into a whole, which helps balance attack performances between different modalities. Here, we use a simple sum of Eq.(\ref{dis_vis}) and Eq.(\ref{dis_inf}) instead of Eq.(\ref{J(s)}) as the not-combined fitness function.
\begin{table}[h]
\vspace{-0.2cm}
\centering
   \caption{Ablation study for score-aware iterative fitness function.}
\begin{tabular}{c |c |c}
  \hline  
        &  Ours & Sum  \\   \hline  
ASR     &       \textbf{ 73.33\% }   &  50.83\%      \\   \hline  
AP drop (Visible)  &  \textbf{     99.19\% }   &  85.37\%  \\   
AP drop (Infrared)   & \textbf{74.31\%}  &  66.99\% \\ \hline  
\end{tabular} 
\vspace{-0.2cm}
\label{combine}
\end{table}

From Table \ref{combine}, we can find that although a simple sum fitness function can have an AP decrease of 85.37\% in the visible modality and 66.99\% in the infrared modality, it suffers a decline in ASR under a cross-modal standard.

\subsubsection{Hyperparameters Tuning}
\begin{figure}[h]
\begin{center}
\includegraphics[width=\linewidth]{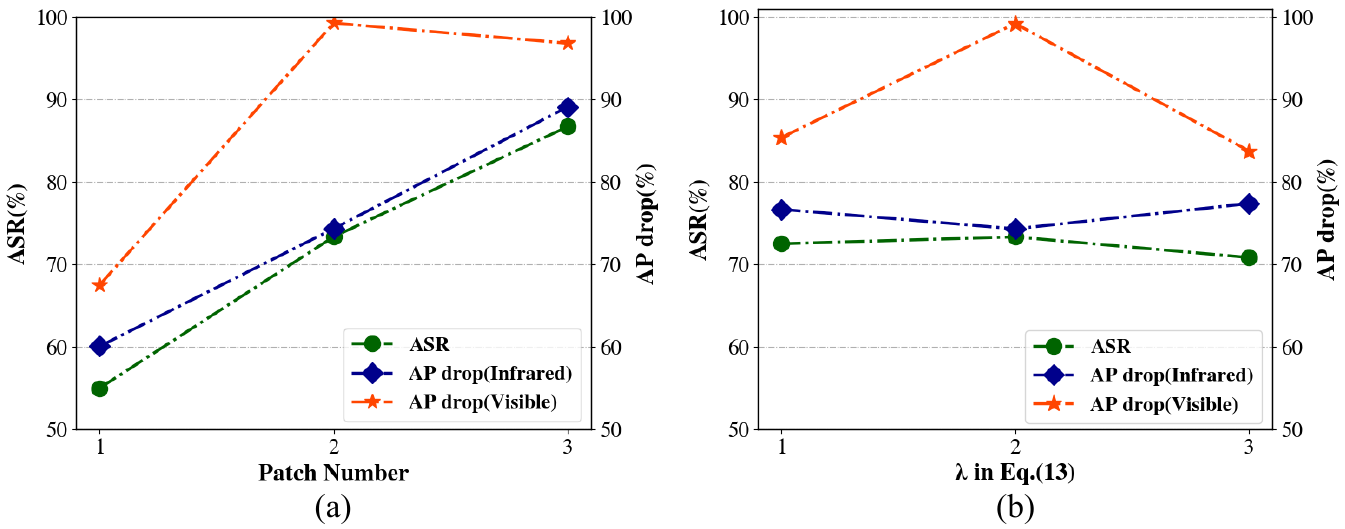}
\end{center}
\vspace{-0.5cm}
\caption{The ASR(\%) and AP drop(\%) of unified adversarial patches
with different patch numbers and hyperparameter $\lambda$.}
\label{fig:patch_number}
\vspace{-0.5cm}
\end{figure}

\textbf{Patch number.} As one of the most directly adjustable hyperparameters, patch number may have an impact on the attack performance. Therefore, we need to explore whether and how it is related to the attack performances. Figure \ref{fig:patch_number} (a) vividly shows that as the number of patches rises, both ASR and AP drop increase with our expectation. However, considering that our attack method has achieved acceptable results with two patches, and three patches would bring larger physical implementation errors, we finally decide to choose the patch number 2 as the main evaluation option.

\noindent\textbf{Hyperparameter $\lambda$.} The $\lambda$ in Eq.(\ref{J(s)}) affects the variability of the final fitness of the individual, the larger the $\lambda$, the greater the difference between different individuals, but this does not mean that the greater the difference, the better the attack effects. As shown in Figure \ref{fig:patch_number} (b), the attack effectiveness of cross-modal patches increases when $\lambda$ equals from 1 to 2, but drops dramatically when $\lambda$ equals 3. As a result,  we set $\lambda=2$ to obtain the optimal effect.

 \begin{figure*}[htp]
\begin{center}
\includegraphics[width=0.91\linewidth]{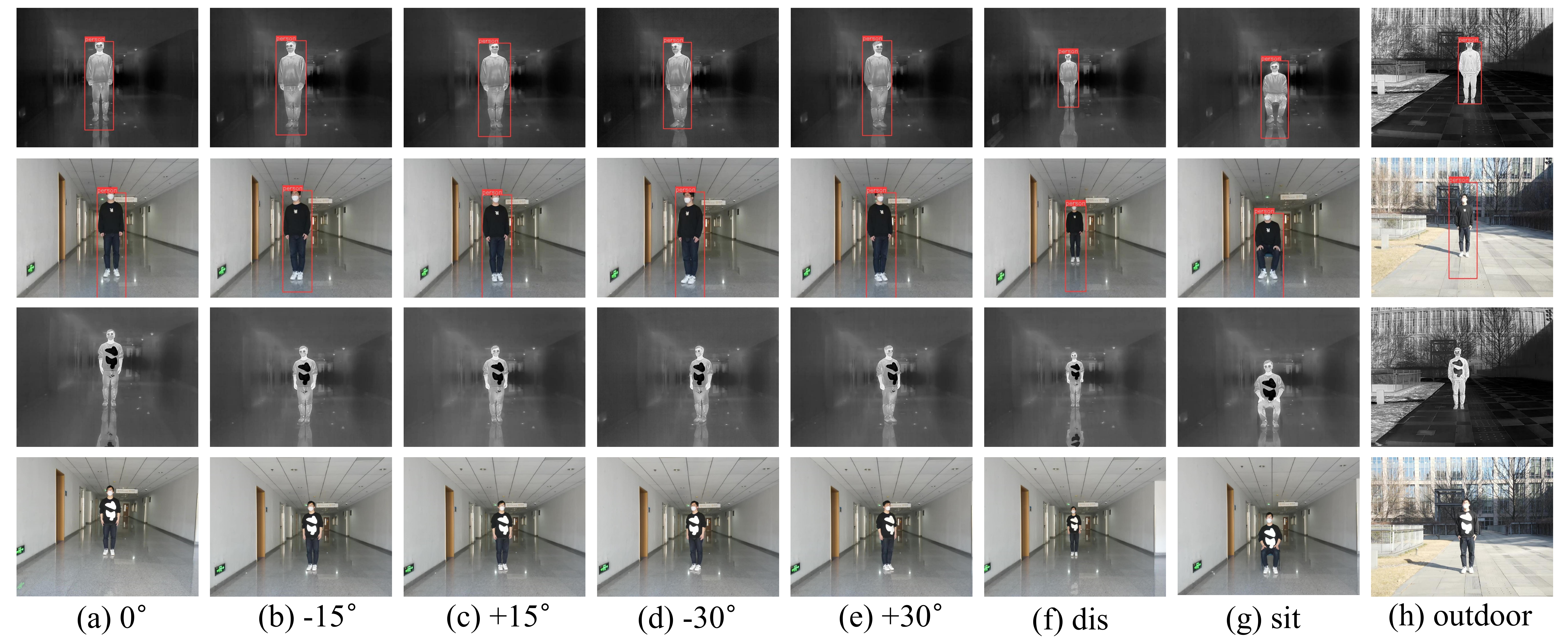}
\end{center}
\vspace{-0.5cm}
\caption{Visual examples of physical attacks with cross-modal patches under various angles, distances, postures, and scenes. }
\label{fig:physical}
\end{figure*}

\subsubsection{Robustness to Implementation Errors}
Since the patch is in general generated digitally by considering the optimal shape and fixed location, when it is applied to real scenarios, it is natural to see how sensitive it would be to the attack success rate if we do not cut and paste the patch in 100\% exact shape and location on the clothes. 

Therefore, we simulate the possible position shifts and clipping errors when doing physical experiments. Table \ref{errors} demonstrates our cross-modal patches' robustness to translation errors and incompleteness.
\begin{table}[!htb]
\caption{ASR of simulating implementation errors.}
\resizebox{\linewidth}{!}
{
\begin{tabular}{c|cc|cc}
\hline
\multirow{2}{*}{} & \multicolumn{2}{c|}{Translation} & \multicolumn{2}{c}{Incompleteness}
\\ \cline{2-5}  & \multicolumn{1}{c|}{3pix}  & 5pix  & \multicolumn{1}{c|}{5\%}  & 10\%  \\ \hline
ASR  & \multicolumn{1}{c|}{\begin{tabular}[c]{@{}c@{}}64.17\%\\ ($\downarrow$9.16\%)\end{tabular}} & \begin{tabular}[c]{@{}c@{}}55.00\%\\ ($\downarrow$18.33\%)\end{tabular} & \multicolumn{1}{c|}{\begin{tabular}[c]{@{}c@{}}68.83\%\\ ($\downarrow$5.00\%)\end{tabular}} & \begin{tabular}[c]{@{}c@{}}60.83\%\\ ($\downarrow$2.50\%)\end{tabular} \\ \hline
\end{tabular}
}
\vspace{-0.4cm}
\label{errors}
\end{table}

\subsection{Attacks in the physical world}
 To test the effectiveness of our method in the real world, we build a series of scenarios and record actual videos under corresponding physical settings to compute their ASR. By default, the recordings of interior scenes are conducted 4 meters away from a standing person in frontal perspective ($0^{\circ}$), and we record each scenario for 20 seconds at a rate of 10 frames per second (about 200 frames in total). Similar to \cite{zhu2022infrared}, we set the threshold of pedestrian detection as 0.7. For different settings, in the angle issue, we use $\pm15^{\circ}$ and $\pm30^{\circ}$ for verification. In the distance issue, we change the camera's default setting from 4 meters to 6 meters. In the posture issue, the pedestrian's posture is replaced by a sitting one. In the scene issue, we switch from the default indoor to the outdoor. Figure \ref{fig:physical} lists the concrete illustrations of various scenarios. Besides, the pedestrian is asked to move the body within the range of $5^{\circ}$ of the current posture to take videos while shooting. Quantitative results are shown in Table \ref{tab:situations}. It is clear that from the frontal perspective, our unified adversarial patches could obtain a high ASR (73.50\%). The ASR still maintains a high value (67.00\% and 46.50\%) when the shooting angle varies. The ASR drops to 71.00\% when the distance is increased to 6 meters from 4 meters. The ASR drops to 61.50\% when the posture is changed from standing to sitting, and to 57.00\% at exterior scenes. These findings demonstrate that our patches are not significantly impacted by various shooting scenarios. In other words, the effects of adversarial attacks can be maintained as long as the camera can capture the whole shape of our cross-modal patches on the object.
 

\begin{table}[!htb]
\caption{ASR in the physical world when changing angles, distances, postures and scenes captured by multi-modal sensors.}
\vspace{-0.5cm}
  \begin{center}
  \resizebox{\linewidth}{!}{
    \begin{tabular}{c|c|c|c|c|c|c}
    \hline
    &  $0^{\circ}$ &  $\pm 15^{\circ}$ & $\pm 30^{\circ}$  & dist. & pos. & outdoor \\
    \hline
    ASR & 73.50\% & 67.00\%  & 46.50\% & 71.00\% & 61.50\% &57.00\%\\
    \hline
    \end{tabular}}
    \label{tab:situations}
  \end{center}
\vspace{-0.8cm}
\end{table}

\subsection{Defenses against  Unified Adversarial Patches}
\label{defense}
To verify the robustness towards adversarial defense, we utilize two common methods in the digital world: spatial smoothing \cite{xu2017feature} as a pre-processing defense and adversarial training \cite{goodfellow2014explaining} as a post-processing defense. The original ASR is 73.33\% and the results in Table \ref{tab:defense1} shows that:(1) ASR only decreases by 9.16\% after spatial smoothing. This is reasonable since our cover image have the same value that could not be altered by smoothing. (2) In the second case, ASR drops by 23.33\%, which is still within an acceptable range. The whole results prove our method's robustness.
\begin{table}[!htb]
\caption{Results against the defense methods.}
  \begin{center}
    \begin{tabular}{c|c|c}
    \hline
    Defense Methods &  ASR  & Error\\
    \hline
    Spatial Smoothing \cite{xu2017feature} & 64.17\% &9.16\%\\
    \hline
    Adversarial Training \cite{goodfellow2014explaining} & 50.00\% & 23.33\%\\
    \hline
    \end{tabular}
    \label{tab:defense1}
  \end{center}
\vspace{-0.8cm}
\end{table}

\section{Conclusion}

In this paper, we propose a unified adversarial patch in the physical world. For that, we uncover the property that can react both in the visible and infrared modalities: \textbf{shape}. Then, combining the boundary-limited shape optimization with the score-aware iterative fitness evaluation, we guarantee an efficient exploration of the adversarial shape and the balance between different modalities. Experiments on the pedestrian detection in the digital world and physical world verify the effectiveness of our method.

{\small
\bibliographystyle{ieee_fullname}
\bibliography{egbib}
}

\end{document}